\documentclass{article}

\usepackage{arxiv}

\usepackage[utf8]{inputenc} 
\usepackage[T1]{fontenc}    
\usepackage{hyperref}       
\usepackage{url}            
\usepackage{booktabs}       
\usepackage{amsfonts}       
\usepackage{nicefrac}       
\usepackage{microtype}      
\usepackage{lipsum}
\usepackage{graphicx}
\graphicspath{ {./images/} }
\usepackage{subcaption} 
\usepackage{float}

\usepackage{braket}

\title{Schmidt Decomposition-Based Methods for Efficient Quantum Image Encoding}

\author{
Ana-Maria Pangeva* \\
  Fraunhofer Institute for Industrial Mathematics ITWM\\
  Technical University of Sofia\\
  1000 Sofia, Bulgaria\\
  \texttt{ana.maria.pangeva@gmail.com} \\
  \textit{Corresponding author}
   \And
 Yassine Ferhi\\
  Fraunhofer Institute for Industrial Mathematics ITWM\\
Institut d'Optique Graduate School - Université Paris-Saclay\\
91120 Palaiseau, France\\
  \texttt{yassine.ferhi@institutoptique.fr} \\
  \And
 Alexander Geng*\\
  Fraunhofer Institute for Industrial Mathematics ITWM\\
67663 Kaiserslautern, Germany\\
  \texttt{alexander.geng@itwm.fraunhofer.de}\\
  \textit{Corresponding author}\\
  \And
 Andreas Weinmann \\
  Algorithms for Computer Vision, Imaging and Data Analysis Lab\\
  Technical University of Applied Sciences\\
  Schweinfurt, Germany 15213 \\
  \texttt{andreas.weinmann@thws.de} \\
  \And
 Desislava Ivanova \\
  Technical University of Sofia\\
  1000 Sofia, Bulgaria \\
  \texttt{d\_ivanova@tu-sofia.bg} \\
     \And
 Ali Moghiseh\\
  Fraunhofer Institute for Industrial Mathematics ITWM\\
67663 Kaiserslautern, Germany\\
  \texttt{ali.moghiseh@itwm.fraunhofer.de} \\
}

\begin{document}
\maketitle
\begin{abstract}
In quantum image processing, a fundamental step is encoding classical image data into quantum states. This can be achieved using methods such as Flexible Representation of Quantum Images (FRQI), Quantum Probability Image Encoding (QPIE), and Novel Enhanced Quantum Representation (NEQR). However, on real quantum hardware, these encodings can quickly lead to circuits with many gates, large circuit depth, and high qubit usage, which is a problem for Noisy Intermediate-Scale Quantum (NISQ) devices. In this work, we investigate whether low-rank state approximation, formulated via Schmidt decomposition, can help reduce this complexity. The method keeps only the most significant parts of a quantum state’s entanglement structure, making state preparation more efficient while preserving most of the image information. We compare the three encoding techniques in their original form and with low-rank approximation, evaluating metrics such as circuit depth, CNOT count, MSE, and visual quality of reconstructed images. The results reveal meaningful trade-offs between accuracy and resource efficiency, with the FRQI model achieving a 97\% reduction in circuit depth while maintaining a near-perfect reconstruction (MSE $\approx$ 0.27). This demonstrates the potential of low-rank techniques for advancing practical quantum image processing on near-term hardware.
\end{abstract}

\maketitle
\section{Introduction}
The escalating volume and complexity of digital data, particularly high-resolution images, pose significant challenges for classical computing systems ~\cite{yan2017handbook,geng2024thesis}. In fields ranging from medical diagnostics to autonomous navigation, the need to process large-scale visual data often creates bottlenecks in storage and analysis, exceeding the capabilities of even powerful supercomputers~\cite{singh2025quantumeigenface} To address these limitations, Quantum Image Processing (QIP) has emerged as a promising alternative that offers a new paradigm for representing and manipulating visual information ~\cite{yan2025lessons,geng2024thesis}

By leveraging the unique properties of quantum mechanics, such as superposition and entanglement, Quantum Image Processing (QIP) can encode an image of $N$ pixels into a quantum state using a logarithmic number of quantum bits (qubits) -  often on the order of $\log_{2}(N)$~\cite{singh2025quantumeigenface, geng2024thesis}. This technique represents an exponential reduction in memory requirements compared to classical methods~\cite{yan2017handbook}. This compact data representation is more than just a storage solution; it is the foundation for quantum algorithms designed to achieve significant speedups in tasks such as edge detection, pattern recognition, and image filtering~\cite{yan2017handbook, geng2024thesis, biamonte2018qml}. The potential for such acceleration stems from the inherent parallelism of quantum computing, which allows for simultaneous calculations on all pixels of an image ~\cite{singh2025quantumeigenface,yan2017handbook}.

However, turning these theoretical ideas into practical implementations is still a major challenge due to the limitations of current quantum hardware~\cite{geng2023frqi,geng2024thesis, singh2025quantumeigenface, yan2025lessons}. 
We are currently in the Noisy Intermediate-Scale Quantum (NISQ) era, characterized by devices with a relatively small number of qubits, imperfect gate operations, limited connectivity, and short coherence times~\cite{preskill2018nisq,singh2025quantumeigenface, geng2024thesis}. 
These constraints make quantum states fragile and prone to decoherence caused by noise and environmental interactions~\cite{park2021robust,preskill2018nisq, singh2025quantumeigenface}. 
As a result, the depth of a quantum circuit (the number of sequential gate layers) becomes a critical resource. 
Most existing QIP algorithms rely on deep and complex circuits to generate highly entangled image states, which makes them impractical to run on today's hardware~\cite{geng2023frqi,geng2024thesis,singh2025quantumeigenface,preskill2018nisq}. 
Recent experimental results confirm this limitation: reliable image reconstruction with current models is typically restricted to very small cases, as noise quickly destroys the quantum information in larger, more entangled systems~\cite{geng2023frqi,geng2024thesis,yan2025lessons}.

To address the challenge of encoding classical images on quantum hardware, several Quantum Image Representation (QIR) models have been developed, each offering different trade-offs between qubit usage, accuracy, and circuit complexity ~\cite{geng2024thesis, yan2025lessons}. 
The \textit{Flexible Representation of Quantum Images} (FRQI)~\cite{le2011frqi}, one of the earliest and most influential approaches, uses phase encoding: each pixel's grayscale value is mapped to the rotation angle of a single ancillary qubit entangled with a set of position qubits~\cite{geng2023frqi,geng2024thesis,yan2025lessons}. 
While FRQI is efficient in its qubit requirements, it suffers from a probabilistic image retrieval process~\cite{geng2023frqi,geng2024thesis,yan2025lessons}. 
Because pixel intensities are stored in quantum amplitudes, a large number of repeated measurements (shots) are needed to estimate them accurately, making FRQI highly sensitive to noise and sampling errors on NISQ devices~\cite{geng2023frqi, geng2024thesis}. 

In contrast, the \textit{Novel Enhanced Quantum Representation} (NEQR) uses basis encoding, storing the binary value of each pixel intensity across multiple qubits~\cite{geng2023frqi,geng2024thesis,zhang2013neqr}. 
This allows deterministic and perfectly accurate image retrieval through direct measurement, which is a major advantage over FRQI~\cite{yan2025lessons,zhang2013neqr}. 
However, NEQR requires a larger number of qubits and deeper circuits composed of multi-controlled gates - operations that are difficult to execute reliably on hardware with limited qubit connectivity~\cite{geng2023frqi,geng2024thesis}. 

A third model, the \textit{Quantum Probability Image Encoding} (QPIE), applies amplitude encoding to represent pixel values directly in the probability amplitudes of basis states~\cite{geng2023frqi,geng2024thesis}. QPIE is the most qubit-efficient of the three and can flexibly handle rectangular image dimensions~\cite{yan2025lessons,geng2024thesis}. 
Its main limitation, however, lies in its normalization requirement, which prevents it from distinguishing between uniform images (e.g., all-black and all-white), and, similar to FRQI, retrieving exact pixel values remains probabilistic due to measurement noise~\cite{geng2023frqi,geng2024thesis}.

While much of the existing research has focused on model-specific improvements - such as developing more efficient gate decompositions like the Multi-Adapted Y Rotation (MARY) for FRQI to reduce CNOT gate counts - or on comparing the baseline performance of these models, a unified strategy for optimizing their overall circuit complexity in the NISQ era remains largely unexplored~\cite{geng2023frqi,geng2024thesis,araujo2024lrsp,yan2025lessons,park2021robust}. 
The high level of entanglement required to represent detailed quantum images is the main factor driving circuit depth and resource consumption across all QIR methods~\cite{araujo2024lrsp,geng2024thesis,horodecki2009entanglement}. 
This shared limitation highlights a key research gap: the need for a general, model-independent technique that can systematically reduce entanglement and circuit complexity, making quantum image encoding more practical and resilient on noisy, resource-constrained hardware~\cite{yan2025lessons,araujo2024lrsp,park2021robust,singh2025quantumeigenface}.

This paper addresses the above challenge by introducing Low-Rank Approximation (LRA) based on the Schmidt decomposition as a unified approach to improving the efficiency of quantum image processing~\cite{araujo2024lrsp,geng2023frqi,zhang2013neqr}. 
The Schmidt decomposition provides a standard way to express a pure state of a bipartite quantum system-such as a quantum image divided into two subsystems-as a sum of orthonormal product states~\cite{araujo2024lrsp,nielsen2010quantum,kaye2007intro}. 
The number of terms in this sum, known as the Schmidt rank ($k$), directly measures the degree of entanglement between the two parts of the system~\cite{araujo2024lrsp,horodecki2009entanglement}. 

Our main hypothesis is that many natural images, which contain strong spatial correlations and redundant patterns, can be represented by quantum states that are effectively compressible in the Schmidt basis~\cite{zhang2013neqr,yan2025lessons}. 
In other words, the essential visual information of an image can often be captured by a relatively small number of dominant Schmidt coefficients, allowing the full quantum state to be approximated by a lower-rank version with minimal loss of detail~\cite{araujo2024lrsp,stewart1993svd}. 
By truncating the Schmidt expansion and retaining only the most significant coefficients, we produce a low-entanglement approximation of the image state~\cite{araujo2024lrsp,zhang2013neqr}. 
This approximation substantially reduces the resources needed for state preparation, since the circuit depth and the number of CNOT gates scale with $\lceil \log_2(k) \rceil$ rather than with the full image dimension~\cite{araujo2024lrsp,geng2023frqi}. 
Furthermore, both theoretical and experimental studies suggest that these low-entanglement states are more robust to noise on NISQ hardware and can, in practice, yield higher-fidelity results than fully entangled “exact” states~\cite{araujo2024lrsp,geng2023frqi,preskill2018nisq}.

The main contributions of this work are fourfold. 
First, we present a systematic comparative evaluation of the FRQI, NEQR, and QPIE models under the framework of Low-Rank Approximation (LRA), generating quantum image states at different Schmidt ranks to examine the general applicability of the method~\cite{yan2025lessons,geng2024thesis,yan2017handbook}. 
Second, we quantify the performance trade-offs using concrete metrics such as circuit depth~\cite{geng2023frqi,geng2024thesis,araujo2024lrsp}, CNOT gate count~\cite{geng2023frqi,araujo2024lrsp, feniou2024sparse}, and Mean Squared Error (MSE)~\cite{geng2024thesis}, providing a detailed analysis of the balance between circuit complexity and reconstruction fidelity~\cite{yan2025lessons,geng2023frqi,araujo2024lrsp}. 
Third, our results reveal a novel phenomenon we refer to as \textit{discrete rank progression}, where image quality does not improve continuously with increasing rank but rather in distinct steps~\cite{araujo2024lrsp}. 
This pattern suggests the existence of specific, resource-efficient ranks that capture the essential image information with minimal overhead~\cite{araujo2024lrsp,daskin2023dimension}. 
Finally, we discuss the broader implications of these findings for NISQ-era quantum computing~\cite{yan2025lessons,geng2024thesis,cross2019validating}, showing that LRA provides a practical, model-independent approach for making quantum image processing more feasible on current and near-term devices~\cite{araujo2024lrsp,west2023drastic}.

The remainder of this paper is organized as follows. 
The \textit{Methodology} section introduces the three quantum image representation models - FRQI, QPIE, and NEQR, and describes the application of Low-Rank Approximation (LRA) within the unified simulation framework. 
The \textit{Results} section presents the experimental findings, including quantitative evaluations of circuit complexity and reconstruction fidelity across varying Schmidt ranks. 
The \textit{Discussion} section interprets these results in the broader context of quantum hardware limitations, emphasizing the implications of low-rank optimization for practical implementation. 
Finally, the \textit{Conclusion} summarizes the main outcomes of this study and outlines potential directions for extending the present work toward scalable quantum image processing on near-term devices.

\section{Methods: Low-Rank Approximation for Efficient Quantum Image Encoding}

The practical application of quantum image processing (QIP) on current hardware is fundamentally limited by the substantial resource requirements of existing encoding protocols~\cite{yan2017handbook,yan2025lessons}. While these methods offer theoretical advantages in data compression, their corresponding quantum circuits are often too deep and complex for Noisy Intermediate-Scale Quantum (NISQ) devices~\cite{preskill2018nisq}.

This section first provides a brief overview of three prominent Quantum Image Representation (QIR) schemes that serve as the foundation for our work. The primary focus is then placed on our proposed optimization strategy: leveraging low-rank approximation via the Schmidt decomposition to significantly reduce the circuit complexity of these encoding methods.

\subsection{Quantum Image Representation Schemes}

To establish a baseline for optimization, we consider three widely studied methods for encoding grayscale images into quantum states.

\begin{enumerate}
  \renewcommand{\labelenumi}{(\arabic{enumi})}

  \item \textbf{Flexible Representation of Quantum Images (FRQI)} \\
 In FRQI, each pixel’s grayscale value is encoded as a rotation angle on a single ‘color’ qubit, which is then entangled with a register of ‘position’ qubits representing spatial coordinates~\cite{yan2017handbook, geng2024thesis}.  
 An image can be represented as

\[
|I(\theta)\rangle =
\frac{1}{2^n}
\sum_{i=0}^{2^{2n}-1}
\left(\cos(\theta_i)|0\rangle + \sin(\theta_i)|1\rangle\right)
\otimes |i\rangle .
\]

where \(\theta_i\) encodes the grayscale intensity of pixel \(i\), \(|i\rangle\) represents the pixel position, and the image size is assumed to be \(2^n \times 2^n\).

 While efficient in its use of qubits, retrieving the image is probabilistic because the intensity information is stored in amplitudes that must be estimated through repeated measurements~\cite{zhang2013neqr}.

\item \textbf{Novel Enhanced Quantum Representation (NEQR)} \\
This method utilizes basis encoding~\cite{zhang2013neqr}. Instead of an angle, NEQR stores a pixel's entire grayscale value directly into the basis state of a sequence of qubits~\cite{zhang2013neqr,yan2017handbook}. 
\[
|I\rangle =
\frac{1}{2^n}
\sum_{i=0}^{2^{2n}-1}
|C_i\rangle \otimes |i\rangle .
\]
where \(|C_i\rangle\) denotes the binary representation of the grayscale value of pixel \(i\), \(|i\rangle\) encodes the pixel position, and \(n\) corresponds to the image dimension for a \(2^n \times 2^n\) image.

  Because different basis states are orthogonal, this allows deterministic and accurate retrieval of pixel values without repeated measurements~\cite{zhang2013neqr}. 
  The trade-off is that this approach is resource-intensive, requiring significantly more qubits and deeper quantum circuits compared to other methods~\cite{geng2023frqi,geng2024thesis}.

  \item \textbf{Quantum Probability Image Encoding (QPIE)} \\
  This method is a direct application of amplitude encoding~\cite{geng2024thesis}.
  A grayscale image can be represented as a normalized quantum state

\[
|I\rangle = \sum_{i=0}^{2^{2n}-1} c_i |i\rangle,
\]

where the amplitudes \(c_i\) are obtained by normalizing the pixel intensities such that

\[
\sum_{i=0}^{2^{2n}-1} |c_i|^2 = 1,
\]

and the image size is assumed to be \(2^n \times 2^n\).

  It is the most qubit-efficient of the three, as it stores the normalized intensity of each pixel directly into the probability amplitudes of the quantum state's basis vectors~\cite{geng2023frqi,yan2017handbook}. 
  The basis states themselves only encode the pixel's position~\cite{yan2017handbook}. 
  Its efficiency comes at a cost: pixel values must be normalized, which can lead to information loss, and image reconstruction is probabilistic, requiring many measurements to approximate the original pixel values~\cite{yan2017handbook}.
\end{enumerate}

\subsection{Low-Rank Approximation via Schmidt Decomposition}

The quantum image encoding methods discussed above require very complex circuits, which makes them difficult to use on current hardware~\cite{geng2023frqi}. To address this, we apply a low-rank approximation (LRA) to the quantum image state using the Schmidt decomposition~\cite{araujo2024lrsp}.

\subsubsection{The Schmidt Decomposition}

The Schmidt decomposition is the quantum analogue of the classical Singular Value Decomposition (SVD) and provides a canonical representation for a pure state in a bipartite quantum system~\cite{araujo2024lrsp,nielsen2010quantum}. For any pure state $\ket{\psi}$ in a composite Hilbert space $\mathcal{H}_A \otimes \mathcal{H}_B$, there exist orthonormal bases $\{\ket{i_A}\}$ for subsystem $A$ and $\{\ket{i_B}\}$ for subsystem $B$ such that $\ket{\psi}$ can be written as:
\[
\ket{\psi} = \sum_{i=1}^{k} \lambda_i \ket{i_A} \ket{i_B},
\]
where the $\lambda_i$ are non-negative real numbers known as the \emph{Schmidt coefficients}, which satisfy the normalization condition
\[
\sum_i \lambda_i^2 = 1.
\]~\cite{nielsen2010quantum}
The number of non-zero coefficients, $k$, is called the \emph{Schmidt rank} of the state~\cite{nielsen2010quantum,horodecki2009entanglement}. The Schmidt rank is a fundamental measure of entanglement: a state is entangled if and only if its Schmidt rank is greater than one~\cite{nielsen2010quantum}. By treating the register of qubits encoding an image as a bipartite system (e.g., by splitting the position qubits into two halves), we can apply this decomposition to any quantum image state.

\subsubsection{Low-Rank Quantum State Approximation}

A low-rank approximation of a quantum state is achieved by truncating the Schmidt decomposition to include only the $r$ most significant terms, where $r < k$~\cite{eckart1936approximation, araujo2024lrsp, nielsen2010quantum}. The approximated state $\ket{\psi^{(r)}}$ is constructed by summing over the first $r$ Schmidt coefficients and then renormalizing the state to ensure unit norm~\cite{araujo2024lrsp}:
\[
\ket{\psi^{(r)}} = \frac{1}{\sqrt{N_r}} \sum_{i=1}^{r} \lambda_i \ket{i_A} \ket{i_B},
\]
where
\[
N_r = \sum_{i=1}^{r} \lambda_i^2.
\]
This procedure creates a compressed representation of the quantum state that preserves its most dominant features ~\cite{araujo2024lrsp}. The error introduced by truncation is directly quantifiable by the sum of the squares of the discarded Schmidt coefficients~\cite{eckart1936approximation},
\[
\epsilon = \sum_{i=r+1}^{k} \lambda_i^2.
\]
This provides a direct and controllable trade-off between the complexity of the quantum state and the accuracy of the final image representation~\cite{araujo2024lrsp}, which we evaluate using Mean Squared Error (MSE) after reconstruction~\cite{geng2024thesis}.

\subsubsection{Circuit Complexity Reduction for NISQ Hardware}

The main motivation for applying Low-Rank Approximation (LRA) in the Noisy Intermediate-Scale Quantum (NISQ) era is to reduce the large amount of resources required for quantum state preparation~\cite{preskill2018nisq}. 
The complexity of constructing a quantum state depends on its entanglement structure, which is measured by the Schmidt rank~\cite{horodecki2009entanglement}. 

Specialized algorithms, such as the Low-Rank State Preparation (LRSP) method~\cite{araujo2024lrsp}, use the Schmidt decomposition to generate quantum circuits that scale with the logarithm of the Schmidt rank $r$, instead of with the full dimension of the state~\cite{shende2006synthesis,mottonen2005transformation}. 
This makes them much more efficient on current hardware.

A smaller Schmidt rank therefore leads to shallower quantum circuits with fewer entangling gates. 
This is especially important for two reasons:
\begin{itemize}
    \item \textbf{Noise mitigation:} Two-qubit gates such as CNOTs are a major source of error on current superconducting and trapped-ion quantum processors~\cite{cross2019validating,kandala2017hardware}. 
    By reducing both the number of CNOTs and the overall circuit depth, LRA lowers the effect of gate errors and decoherence, allowing more reliable execution of quantum algorithms.
    
    \item \textbf{Improved performance on noisy hardware:} On NISQ devices, the approximation error introduced by truncating the Schmidt decomposition is often smaller than the hardware noise that arises when preparing the full, high-rank state~\cite{araujo2024lrsp}. 
    Experiments have shown that low-rank states can achieve results closer to ideal theoretical predictions than exact states prepared on real hardware~\cite{geng2024thesis}. 
    In quantum image processing, this means that reconstructed images from approximated states can have lower MSE and better visual quality than those generated from ``exact'' states on noisy processors. 
    This makes LRA a practical method for achieving quantum advantage on near-term devices~\cite{preskill2018nisq,yan2025lessons}.
\end{itemize}

\subsubsection{Implementation Details and Simulation Environment}

All experiments were conducted in Python~3.9.21 within \textit{Jupyter Notebook}, using \textit{Qiskit}~2.1.1 and the \textit{qclib} library. 
Quantum circuits were executed on the \textit{Qiskit Aer} simulator, while the \textit{IBM Quantum} platform was used solely for simulator access, without hardware execution.

Low-rank quantum states were generated for all three encoding schemes (\textbf{FRQI}, \textbf{QPIE}, and \textbf{NEQR}) using the \texttt{LowRankInitialize} class based on the Schmidt decomposition. 
For FRQI and QPIE, truncation parameters were varied within $r \in \{1,\dots,64\}$, while NEQR, due to its larger qubit register, supported higher ranks up to $2^{10}$. 
For each configuration, circuits were transpiled on the \textit{AerSimulator}, and their circuit depth and two-qubit (CNOT) gate counts were recorded. 
Reconstructed states were measured and compared to their classical counterparts using the Mean Squared Error (MSE) metric to quantify reconstruction fidelity.

All simulations were performed on an Intel(R) Xeon(R) Gold~6442Y workstation with 500\,GB RAM, running Ubuntu~Linux~(64-bit). 
The supporting Python libraries included NumPy~2.0.2, Matplotlib~3.9.2, Scikit-image~0.24.0, Pillow~11.3.0, and Pandas~2.2.2.

\section{Results}

\begin{figure}[h]
    \centering
    \includegraphics[width=0.20\textwidth]{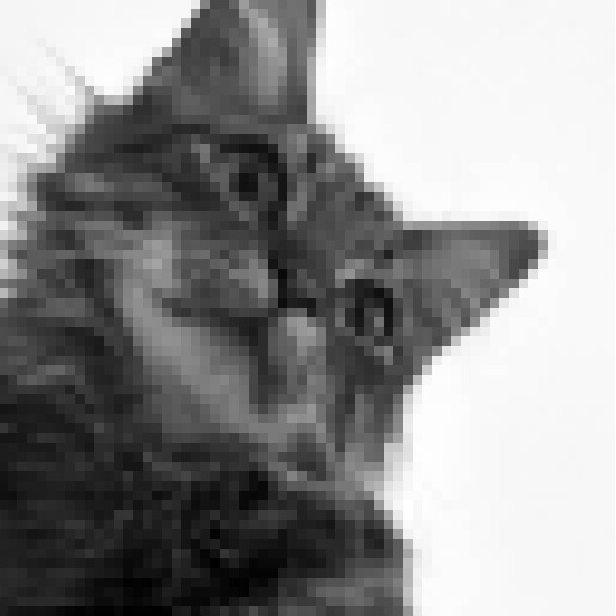}
    \caption{64×64 grayscale input image used for all encoding experiments.}
    \label{fig:input_image}
\end{figure}

Figure~\ref{fig:input_image} shows the 64×64 grayscale image used as the input dataset for all experiments. 
This image was chosen to balance visual complexity with simulation feasibility, providing enough detail to evaluate reconstruction quality under different encoding schemes. 
All encodings were applied using the experimental setup described in the previous section, and results were analyzed in terms of circuit depth, CNOT count, and Mean Squared Error (MSE).

\subsection{Flexible Representation of Quantum Images (FRQI).}

The FRQI method used 13 qubits - 12 for pixel position and one for intensity encoding. 
The full-rank circuit ($r = 64$) reached a depth of 385{,}025 and 221{,}184 CNOT gates. 
After applying low-rank approximation, these values decreased to 11{,}256 and 7{,}649 at rank $r = 33$, 
corresponding to a reduction of approximately 97\% in both metrics.

\begin{figure}[h]
  \centering
  \setlength{\tabcolsep}{1.5pt} 
  \renewcommand{\arraystretch}{1.5} 

  \begin{tabular}{cc}
    \includegraphics[width=.17\textwidth]{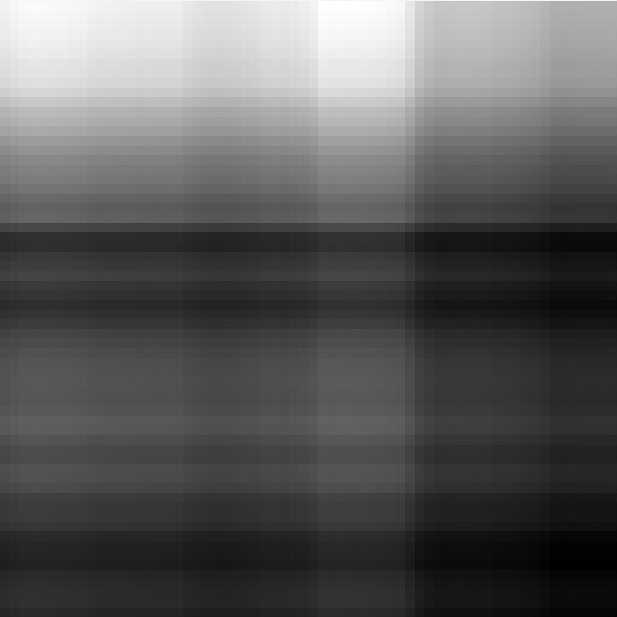} &
    \includegraphics[width=.17\textwidth]{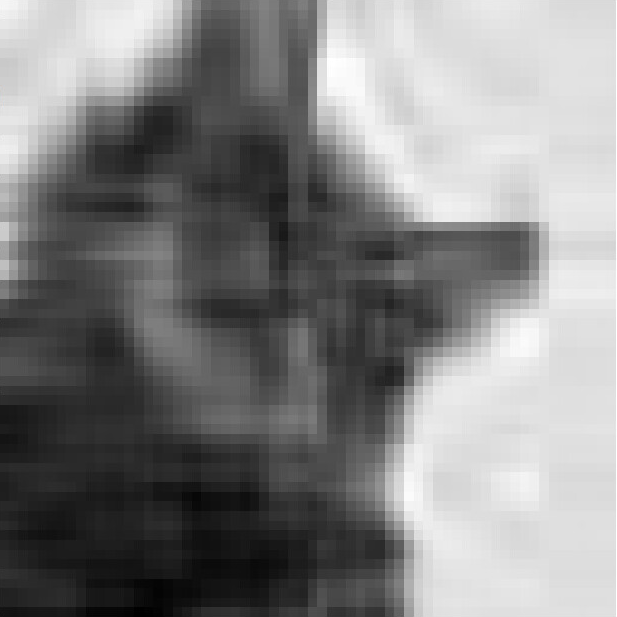} \\[-2pt]
    \footnotesize (a) $r=1$ & \footnotesize (b) $r=5$ \\[2pt]
    \includegraphics[width=.17\textwidth]{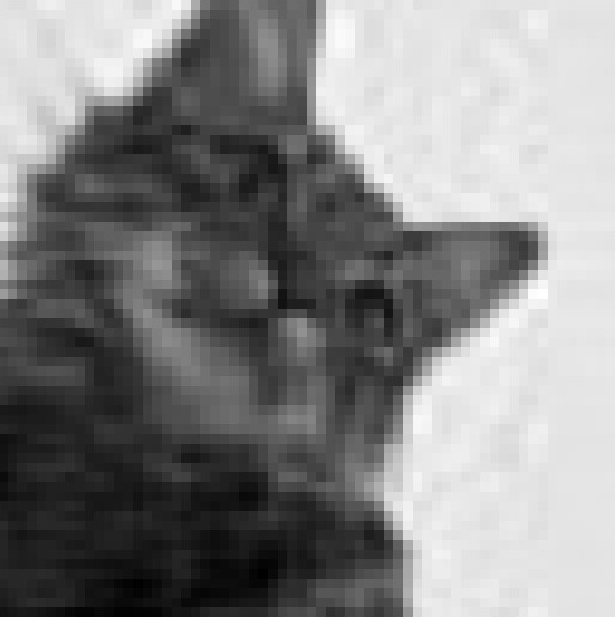} &
    \includegraphics[width=.17\textwidth]{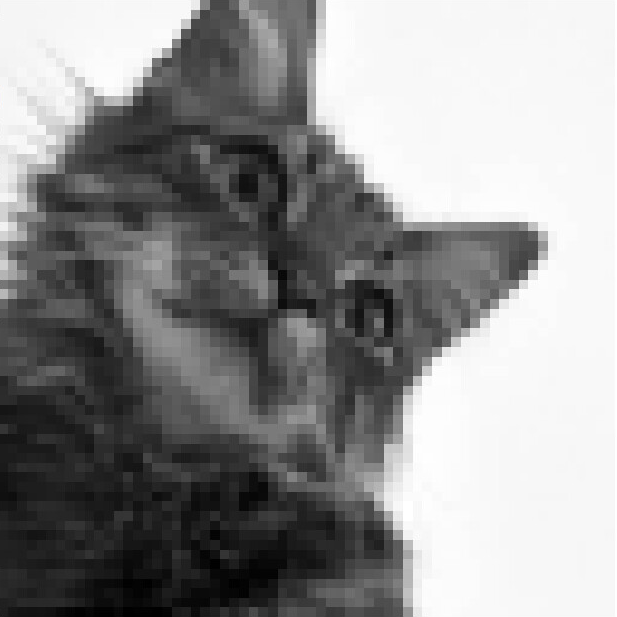} \\[-2pt]
    \footnotesize (c) $r=9$ & \footnotesize (d) $r=33$
  \end{tabular}

  \caption{FRQI reconstructions for increasing Schmidt ranks $r$.}
  \label{fig:frqi_results}
\end{figure}

Despite this drastic simplification, the reconstructed image retained near-perfect fidelity (MSE~$\approx$~0.277). Throughout this work, grayscale values are represented in the range [0,255], meaning that an MSE of approximately 0.277 corresponds to an average pixel intensity error below one grayscale level, which is visually imperceptible. 
As illustrated in Fig.~\ref{fig:frqi_results}, ranks below 5 captured only blurred silhouettes, 
rank~9 recovered most structural features, and rank~33 achieved a visually indistinguishable reconstruction from the original image.

\subsection{Quantum Probability Image Encoding (QPIE).}
\begin{figure}[h]
  \centering
  \setlength{\tabcolsep}{1.5pt} 
  \renewcommand{\arraystretch}{1.5} 

  \begin{tabular}{cc}
    \includegraphics[width=.17\textwidth]{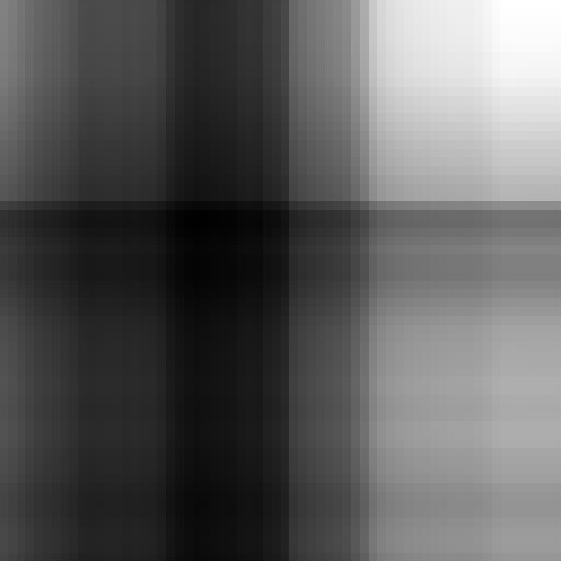} &
    \includegraphics[width=.17\textwidth]{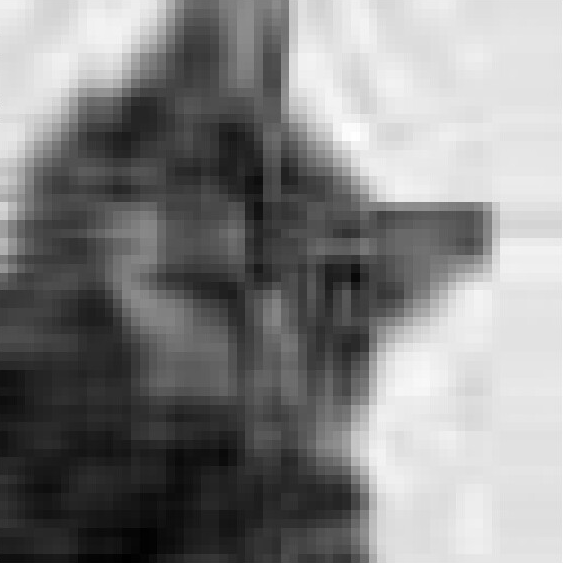} \\[-2pt]
    \footnotesize (a) $r=1$ & \footnotesize (b) $r=5$ \\[2pt]
    \includegraphics[width=.17\textwidth]{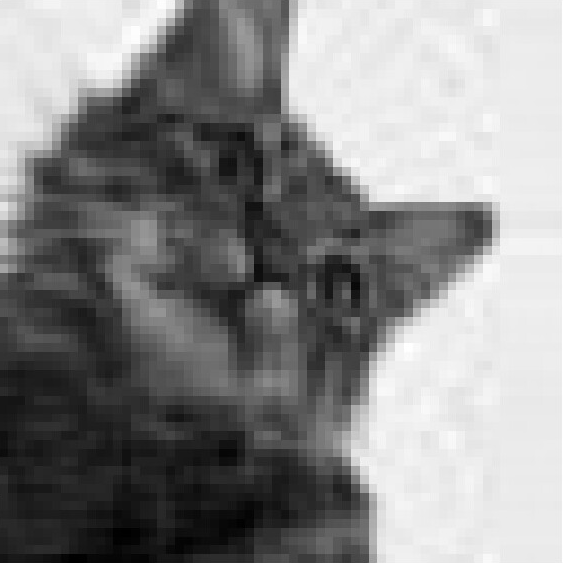} &
    \includegraphics[width=.17\textwidth]{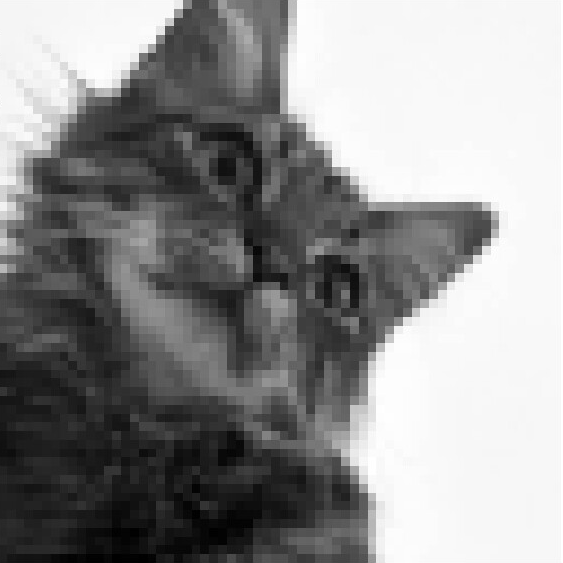} \\[-2pt]
    \footnotesize (c) $r=9$ & \footnotesize (d) $r=33$
  \end{tabular}

  \caption{QPIE reconstructions for increasing Schmidt ranks $r$.}
  \label{fig:qpie_results}
\end{figure}

QPIE utilized only 12 qubits, making it the most compact among the compared representations. 
The full-rank configuration ($r = 64$) produced a circuit of depth 19{,}910 with 4{,}083 CNOT gates. 
Applying the low-rank approximation (LRA) led to a depth of 3{,}704 and 3{,}788 CNOT gates at $r = 33$, corresponding to reductions of 81\% and 7.2\%, respectively. 
The reconstructed image maintained near-perfect quality (MSE~$\approx$~0.272) and was visually indistinguishable from the input, as shown in Fig.~\ref{fig:qpie_results}.

However, the practical cost of QPIE became evident during simulation. 
While its design is elegant and resource-efficient in theory, transpiling the full-rank circuit required nearly three days of computation. 
This exposes a limitation of QPIE on current simulators and NISQ hardware-its excellent encoding compactness comes at the expense of significant compilation time and limited scalability.

\subsection{Novel Enhanced Quantum Representation (NEQR).}

\begin{figure}[h]
  \centering
  \setlength{\tabcolsep}{1.5pt}
  \renewcommand{\arraystretch}{1.5}

  \begin{tabular}{ccc}
    \includegraphics[width=.17\textwidth]{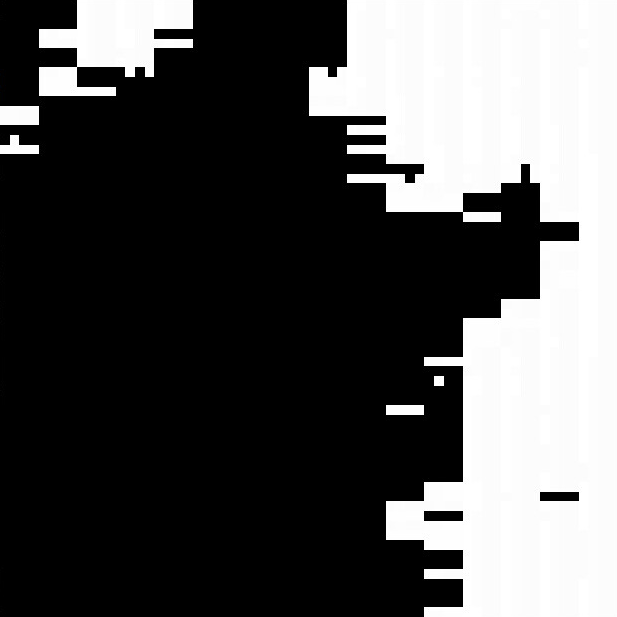} &
    \includegraphics[width=.17\textwidth]{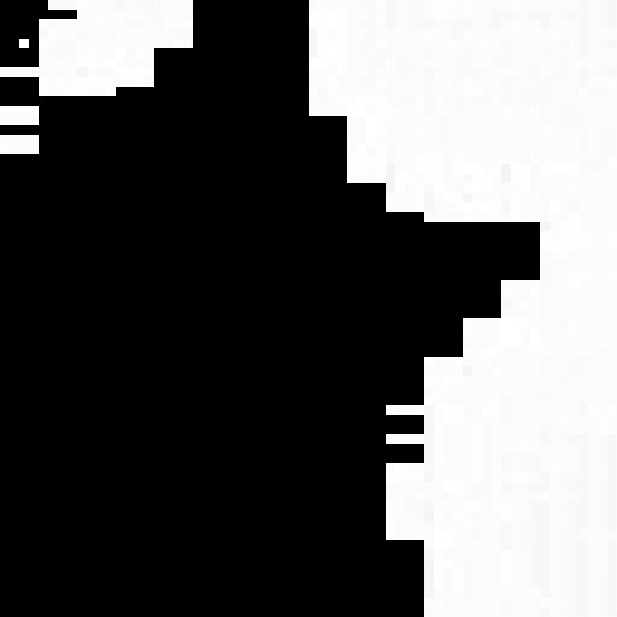} &
    \includegraphics[width=.17\textwidth]{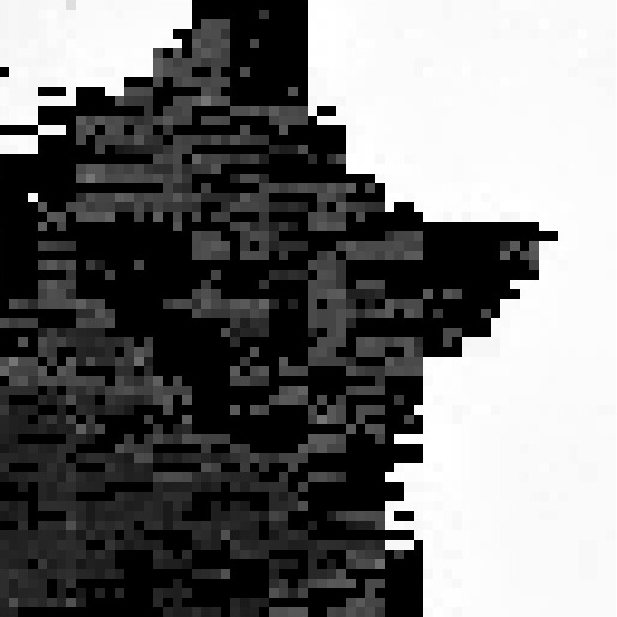} \\[-2pt]

    \footnotesize (a) $r=1$ &
    \footnotesize (b) $r=5$ &
    \footnotesize (c) $r=17$ \\[2pt]

    \includegraphics[width=.17\textwidth]{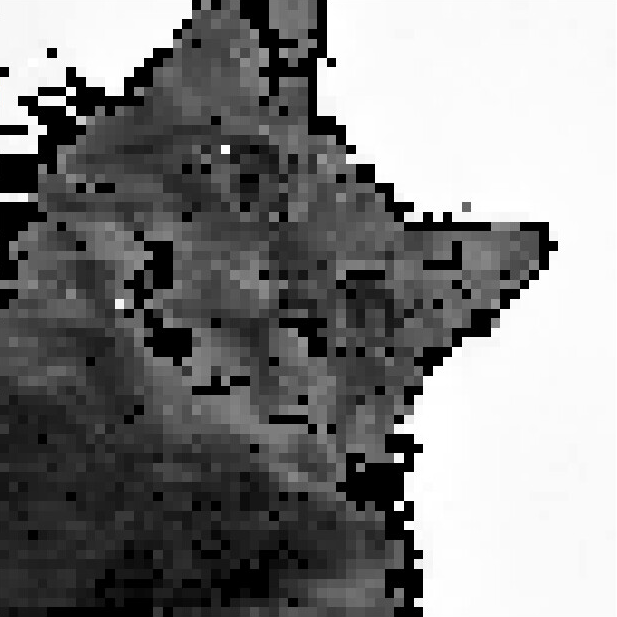} &
    \includegraphics[width=.17\textwidth]{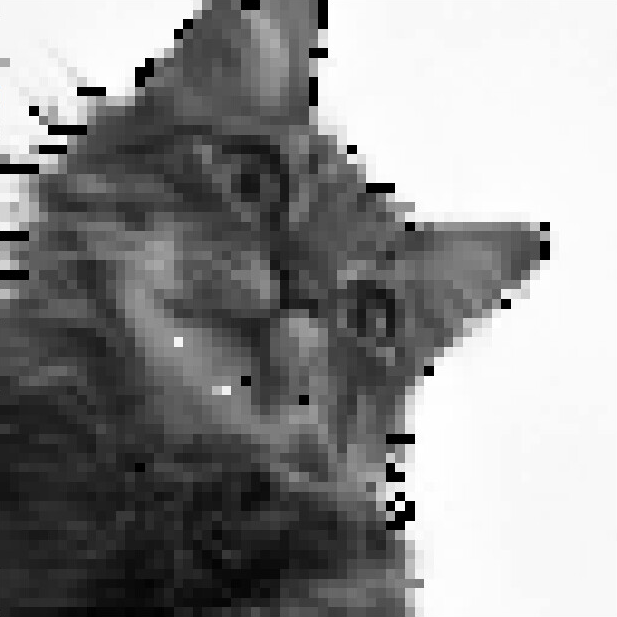} &
    \includegraphics[width=.17\textwidth]{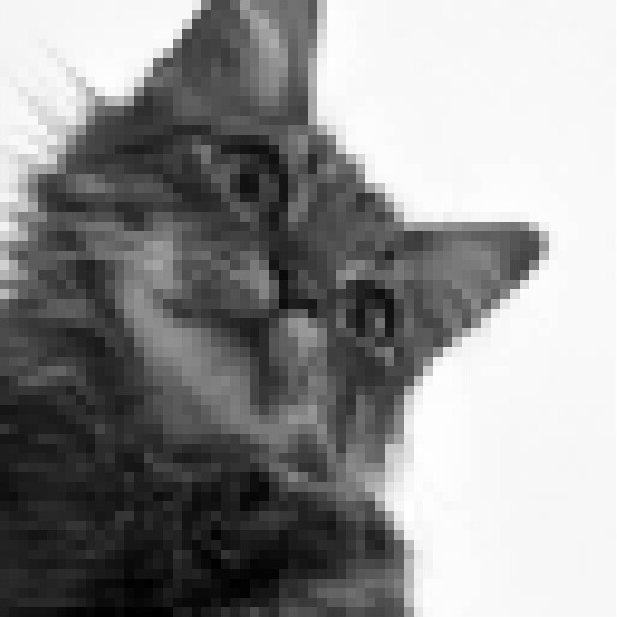} \\[-2pt]

    \footnotesize (d) $r=65$ &
    \footnotesize (e) $r=129$ &
    \footnotesize (f) $r=257$
  \end{tabular}

  \caption{NEQR reconstructions for increasing Schmidt ranks $r$.}
  \label{fig:neqr_results}
\end{figure}
NEQR used 20 qubits - 12 to locate each pixel and 8 to encode its grayscale intensity, making it the most demanding scheme in terms of qubit count and circuit size. 
The full-rank configuration ($r = 1024$) reached a depth of 2{,}737{,}166 with 2{,}072{,}898 CNOT operations.

After applying the low-rank approximation (LRA), these values dropped sharply: 
at rank $r = 257$, the circuit depth was reduced to 735{,}168 (–73\%) and the number of CNOTs to 751{,}554 (–64\%), with a reconstruction error of $\mathrm{MSE}=0.2728$. 
At $r = 513$, the image was recovered perfectly ($\mathrm{MSE}=0$), showing that most visual information is concentrated in the dominant Schmidt components.

Figure~\ref{fig:neqr_results} illustrates the reconstruction process. 
At low ranks ($r=1$–$5$), only scattered black regions appear, representing isolated intensity clusters. 
From $r=17$ to $r=65$, the outline of the object emerges, though the distribution remains blocky due to discrete grayscale encoding. 
Ranks $r=129$ and $r=257$ progressively restore texture and contrast until the image becomes virtually identical to the original. 
Compared with FRQI and QPIE, NEQR achieves pixel-level accuracy but at the cost of extremely deep and gate-heavy circuits, limiting its feasibility on near-term hardware.

\begin{figure}[h]
  \centering
  \includegraphics[width=0.6\textwidth]{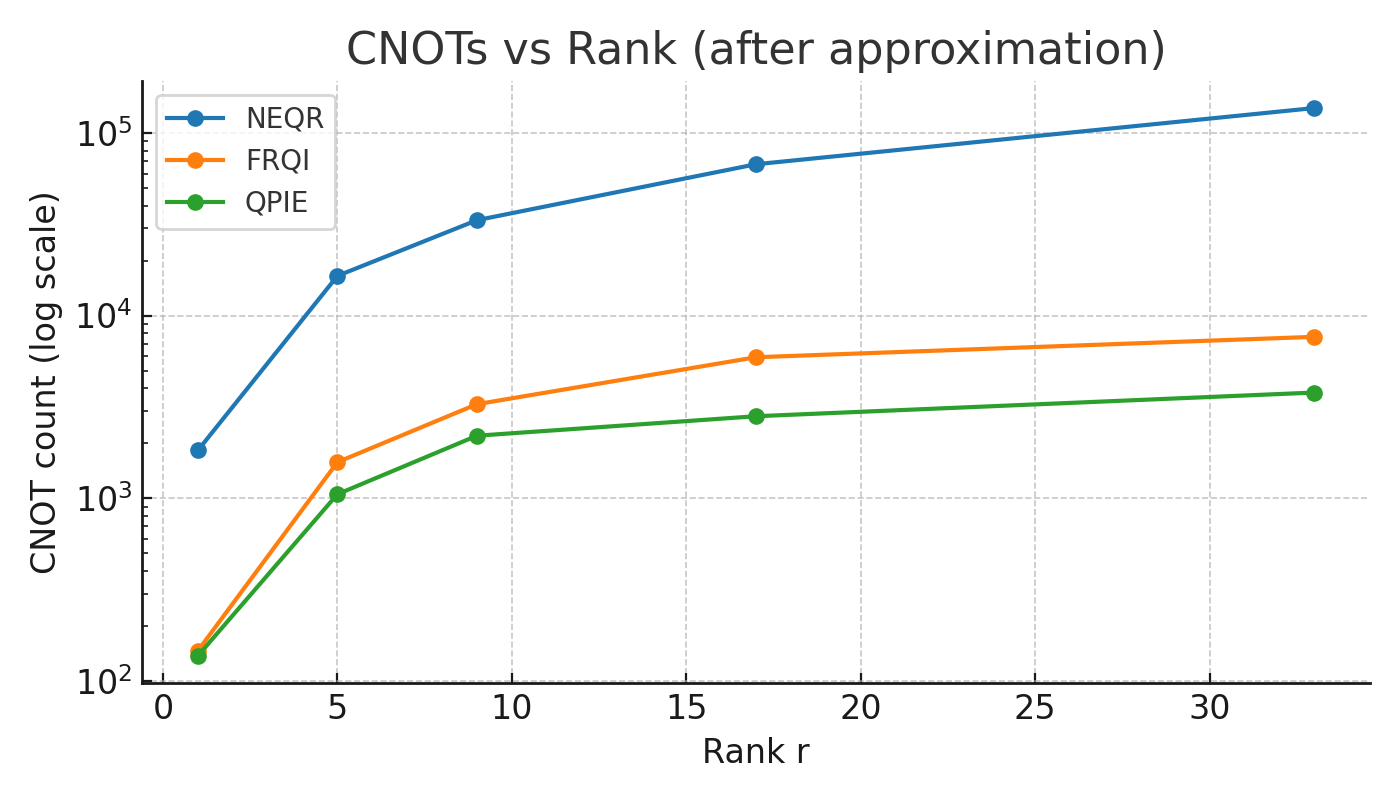}
  \caption{
  CNOT count vs.\ Schmidt rank (log scale) after applying Low-Rank Approximation (LRA).}
  \label{fig:cx_rank}
\end{figure}

\begin{figure}[h]
  \centering
  \includegraphics[width=0.6\textwidth]{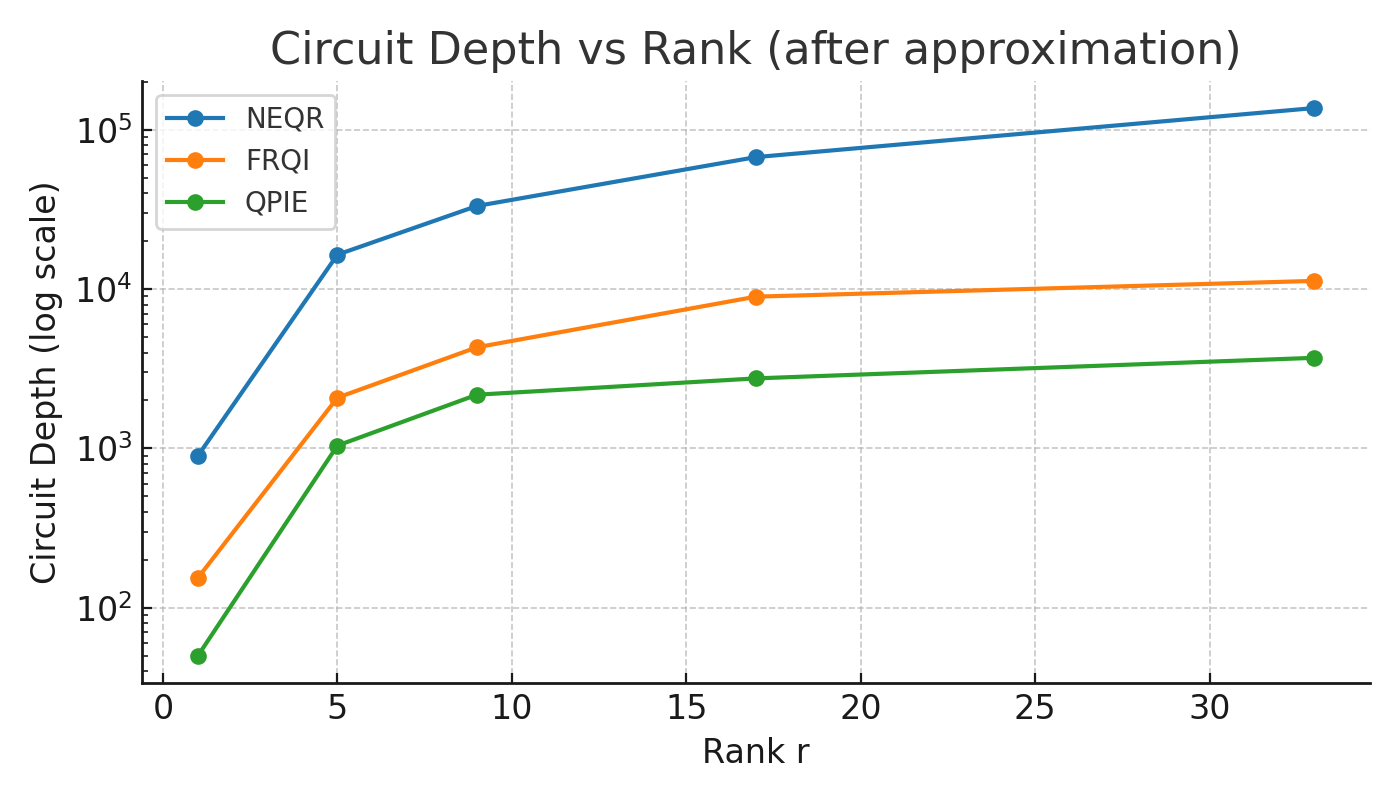}
  \caption{
  Circuit depth vs.\ Schmidt rank (log scale). 
  }
  \label{fig:depth_rank}
\end{figure}

\begin{figure}[tb]
  \centering
  \includegraphics[width=0.60\textwidth]{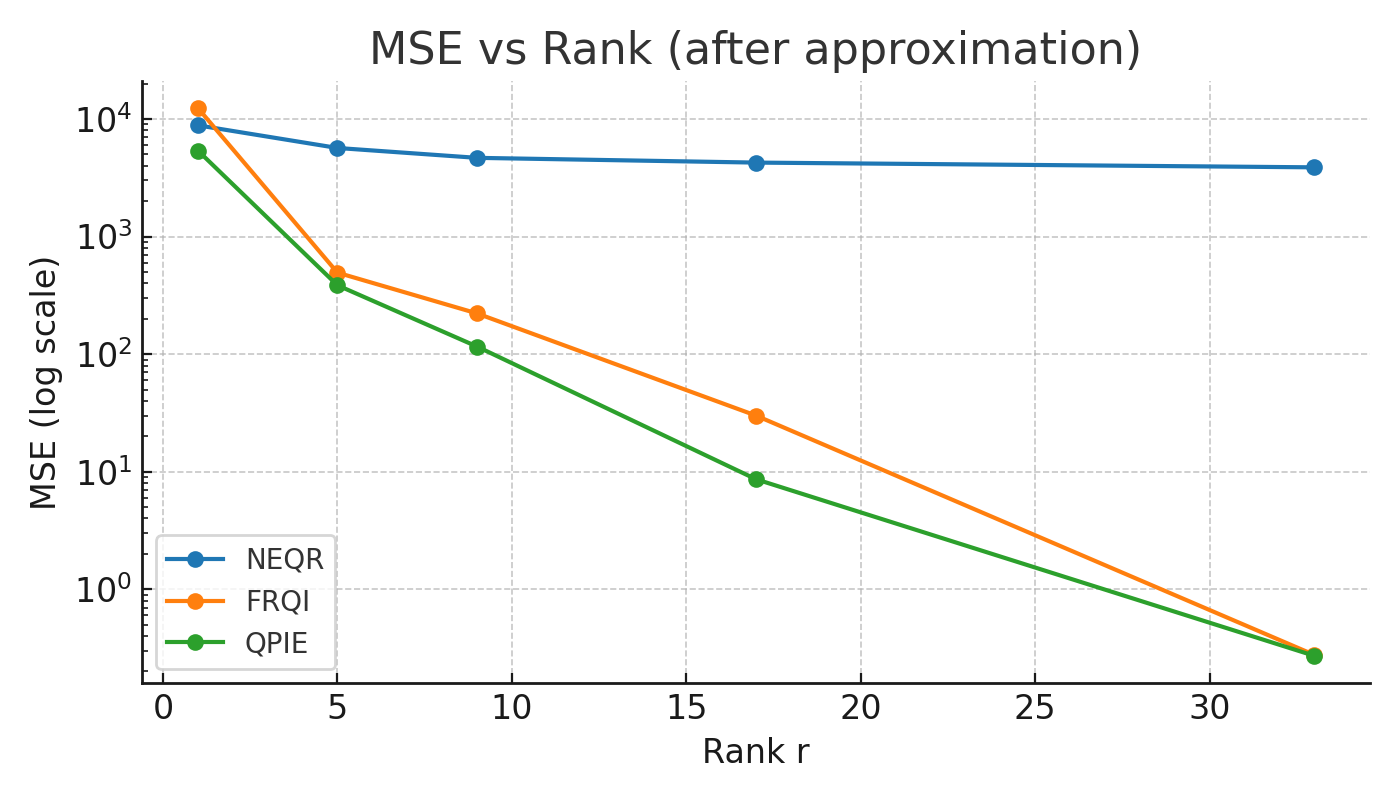}
  \caption{
  Mean Squared Error (MSE) vs.\ Schmidt rank (log scale). 
  }
  \label{fig:mse_rank}
\end{figure}

To complement the visual reconstructions, Figs.~\ref{fig:cx_rank}--\ref{fig:mse_rank} summarize the quantitative behavior of all three encoding schemes after applying Low-Rank Approximation (LRA). 
As shown in Fig.~\ref{fig:cx_rank}, the number of CNOT gates grows sublinearly with the Schmidt rank, confirming that circuit complexity remains well-controlled even for higher ranks. 
A similar trend is observed in Fig.~\ref{fig:depth_rank}, where circuit depth follows the same scaling pattern, highlighting the efficiency of LRA in producing shallower circuits suitable for NISQ-era devices. 
Finally, Fig.~\ref{fig:mse_rank} illustrates the corresponding Mean Squared Error (MSE), which decreases exponentially with increasing rank-indicating that higher ranks recover more visual detail while maintaining manageable circuit sizes. 
Together, these plots demonstrate that low-rank truncation yields a highly favorable balance between circuit efficiency and image fidelity across all tested quantum image representations.

\noindent

Across all three encoding methods, increasing the Schmidt rank improved reconstruction accuracy at the cost of higher circuit complexity. 
FRQI achieved the most substantial optimization, with a 97\% reduction in circuit depth and CNOT count at rank $r=33$ while maintaining excellent fidelity. 
QPIE produced the smallest circuits and required the fewest qubits, although its entanglement reduction was more modest. 
NEQR also benefited from low-rank truncation but remained the most resource-demanding due to its large register size.
Table~\ref{tab:comparison} summarizes the key numerical results for all three encoding methods before and after applying low-rank approximation.

\begin{table}[tb]
\centering
\caption{Comparison of quantum image encoding methods before and after low-rank approximation (LRA) for a $64\times64$ grayscale image.}
\label{tab:comparison}
\resizebox{0.7\textwidth}{!}{
\begin{tabular}{lcccc}
\toprule
\textbf{Method} & \textbf{Qubits} & \textbf{Depth (Full / LRA)} & \textbf{CNOTs (Full / LRA)} & \textbf{MSE} \\
\midrule
FRQI & 13 & 385{,}025 / 11{,}256 & 221{,}184 / 7{,}649 & 0.277 \\
QPIE & 12 & 19{,}910 / 3{,}704 & 4{,}083 / 3{,}788 & 0.272 \\
NEQR & 20 & 2{,}737{,}166 / 735{,}168 & 2{,}072{,}898 / 751{,}554 & 0.273 \\
\bottomrule
\end{tabular}
}
\end{table}

A consistent progression pattern was observed across all models: noticeable updates in both circuit structure and reconstructed image quality appeared only at specific ranks $r = 1, 2, 3, 5, 9, 17, 33, 65, 129, 257, \ldots$. 
For ranks between these values, the reconstructions and corresponding circuits remained effectively identical to those of the preceding configuration. 
This stepwise behavior was consistently observed across all encoding schemes, 
showing that increasing the rank does not gradually improve the results. 
Instead, new information becomes visible only at specific ranks, 
suggesting that the contribution of additional Schmidt components occurs in discrete steps.

\section{Discussion}
\subsection{Interpretation of Findings}

The experimental results reveal distinct behaviors across the three quantum image encoding schemes under low-rank approximation (LRA). 
FRQI exhibited the highest degree of compressibility, achieving near-perfect reconstruction at $r=33$ while reducing circuit depth and CNOT count by nearly 97\%. 
This indicates that most of the visual information in the FRQI state is contained within a small number of Schmidt components. 
QPIE, although the most qubit-efficient method, showed limited improvement in CNOT reduction and required significantly longer transpilation time, reflecting the computational overhead of amplitude-based state preparation. 
NEQR demonstrated clear benefits from LRA as well, yet remained the most resource-demanding due to its large qubit register and deep circuit structure. 
Overall, the findings suggest that the majority of image information can be encoded and reconstructed using a fraction of the full-state rank, highlighting the potential of LRA as a practical optimization for quantum image representation.

\subsection{Implications for NISQ Hardware}

These findings carry practical implications for NISQ-era quantum computing. 
By reducing circuit depth and the number of entangling gates, LRA mitigates two of the most critical constraints on current hardware-limited coherence times and CNOT gate noise. 
While our experiments were performed in simulation, the observed reductions indicate that shallow, approximate encodings could perform more reliably on real devices than their full-rank equivalents. 
This supports the idea that exact state preparation is not always optimal on noisy hardware: an approximate, low-rank circuit may yield higher effective fidelity after execution due to its reduced susceptibility to noise. 
Thus, LRA can be viewed as a practical design tool for tailoring quantum image algorithms to the capabilities of near-term devices.

\subsection{Limitations and Future Work}

This study was conducted entirely on a noiseless simulator, which isolates the effect of Low-Rank Approximation (LRA) but does not capture hardware-induced errors. 
Future work should therefore validate these results on real quantum processors, where noise and decoherence would provide a more realistic measure of performance. 

Another limitation lies in the dataset, as the experiments were performed on a single $64\times64$ grayscale image. Applying the same methodology to larger and more diverse images would help determine how the Schmidt spectrum and compression efficiency scale with image resolution and structural complexity.

An additional open question concerns the consistent progression pattern observed across all experiments, where meaningful changes appeared only at specific Schmidt ranks $r = 1, 2, 3, 5, 9, 17, 33, 65, \ldots$. 
The origin of this stepwise behavior remains unclear and should be further examined to better understand its mathematical basis and its connection to the structure of quantum image encodings. 
Developing a theoretical explanation for this phenomenon could improve our understanding of how entanglement evolves in quantum image states and guide more principled rank-selection strategies.

Another limitation is that many quantum image processing algorithms rely on the original structure of the encoded image state. After applying Low-Rank Approximation (LRA), this structure is no longer preserved exactly, which may affect how some downstream QIP operations behave. Future work could explore how existing algorithms can be adapted to work directly with low-rank states, or how LRA can be incorporated into the full processing pipeline rather than used only during state preparation.

Finally, combining LRA with other quantum circuit optimization techniques could offer further improvements in efficiency and scalability for future implementations on real NISQ hardware.

\section{Conclusion}

This study investigated the application of Low-Rank Approximation (LRA) based on the Schmidt decomposition as a strategy for reducing the circuit complexity of quantum image encoding. 
By systematically applying LRA to three major quantum image representation models - FRQI, QPIE, and NEQR, we demonstrated that quantum image states can be efficiently approximated by retaining only the most significant Schmidt coefficients. 
This truncation effectively reduces entanglement while preserving the key visual information contained in the original image, resulting in circuits that are both shallower and less error-prone. 
Across all models, the method achieved up to 97\% reduction in circuit depth and 96\% reduction in CNOT count while maintaining a reconstruction error near $\mathrm{MSE} \approx 0.27$, confirming that low-rank truncation provides a favorable balance between fidelity and efficiency on NISQ-era simulators.

The results reveal that the majority of visual information is concentrated within a small subset of Schmidt components, meaning that exact state preparation is often unnecessary for accurate image recovery. 
This finding has important implications for the design of future quantum algorithms, suggesting that approximate state preparation, guided by entanglement structure, may outperform exact methods under realistic hardware constraints. 
Among the three models, FRQI benefited most from the approximation, achieving substantial compression with minimal quality loss, while QPIE demonstrated excellent qubit efficiency but remained computationally demanding during compilation. 
NEQR, although the most resource-intensive, also exhibited significant gains from LRA, showing that even highly entangled binary encodings can be optimized through rank truncation.

An interesting and consistent pattern was observed in all experiments: meaningful updates in both circuit structure and reconstructed image quality appeared only at specific Schmidt ranks ($r = 1, 2, 3, 5, 9, 17, 33, \ldots$). 
This stepwise progression indicates that new entangled components contribute discretely rather than continuously as the rank increases. 
While the exact theoretical explanation for this phenomenon remains open, it likely reflects structural properties of quantum image states and warrants further analytical study.

Future work should extend the current evaluation to larger and more diverse datasets to explore how the Schmidt spectrum scales with image complexity and resolution. Running the optimized low-rank circuits on real quantum hardware will also be essential for assessing their resilience to noise and decoherence. 
A deeper theoretical investigation into the observed discrete rank behavior could lead to more principled strategies for selecting optimal truncation thresholds and further improving circuit synthesis methods. An additional future direction is exploring how Low-Rank Approximation can be integrated into complete quantum image processing pipelines, rather than being used only during state preparation. Together, these directions point toward a broader framework in which low-rank techniques form the foundation for scalable, hardware-aware quantum image processing.

\section*{Acknowledgements}

This work was supported by the Quantum Initiative Rhineland-Palatinate (QUIP) and by the European Union, Interreg Upper Rhine Valley programme within the UpQuantVal project.

\bibliographystyle{unsrt} 
\bibliography{references}  

\end{document}